\documentclass[letterpaper, 10 pt, conference]{ieeeconf}  

\IEEEoverridecommandlockouts                              

\usepackage{booktabs} 
\usepackage{array}    
\usepackage{rotating} 
\usepackage{caption}  
\usepackage{graphicx} 
\usepackage{makecell} 
\usepackage{subcaption} 
\usepackage{multirow}
\usepackage{adjustbox}
\usepackage{amsmath}
\usepackage{amssymb}
\usepackage{siunitx}  
\usepackage{threeparttable} 
\usepackage{xcolor}
\usepackage{cuted}
\usepackage{pifont}
\usepackage{url}
\title{\LARGE \bf
On-the-Fly3R: Towards Robust Online 3D Reconstruction with Feed-Forward 3R Models for Large-Scale UAV Scenarios 
}

\author{Zhe Shen$^{*}$, Liyuan Lou$^{*}$, Yifei Yu, Guanbo Wang, Quanjian Ji, Xin Wang$^{\dagger}$, Zongqian Zhan$^{\dagger}$
\\
School of Geodesy and Geomatics, Wuhan University, Wuhan 430079, China
\thanks{$^{*}$Equal contribution.}
\thanks{$^{\dagger}$Corresponding authors. (xwang, zqzhan)@sgg.whu.edu.cn.}
}

\begin{document}

\maketitle
\thispagestyle{empty}
\pagestyle{empty}
\vspace*{-2cm}  
\begin{strip}
    \centering
    \includegraphics[width=1.0\textwidth]{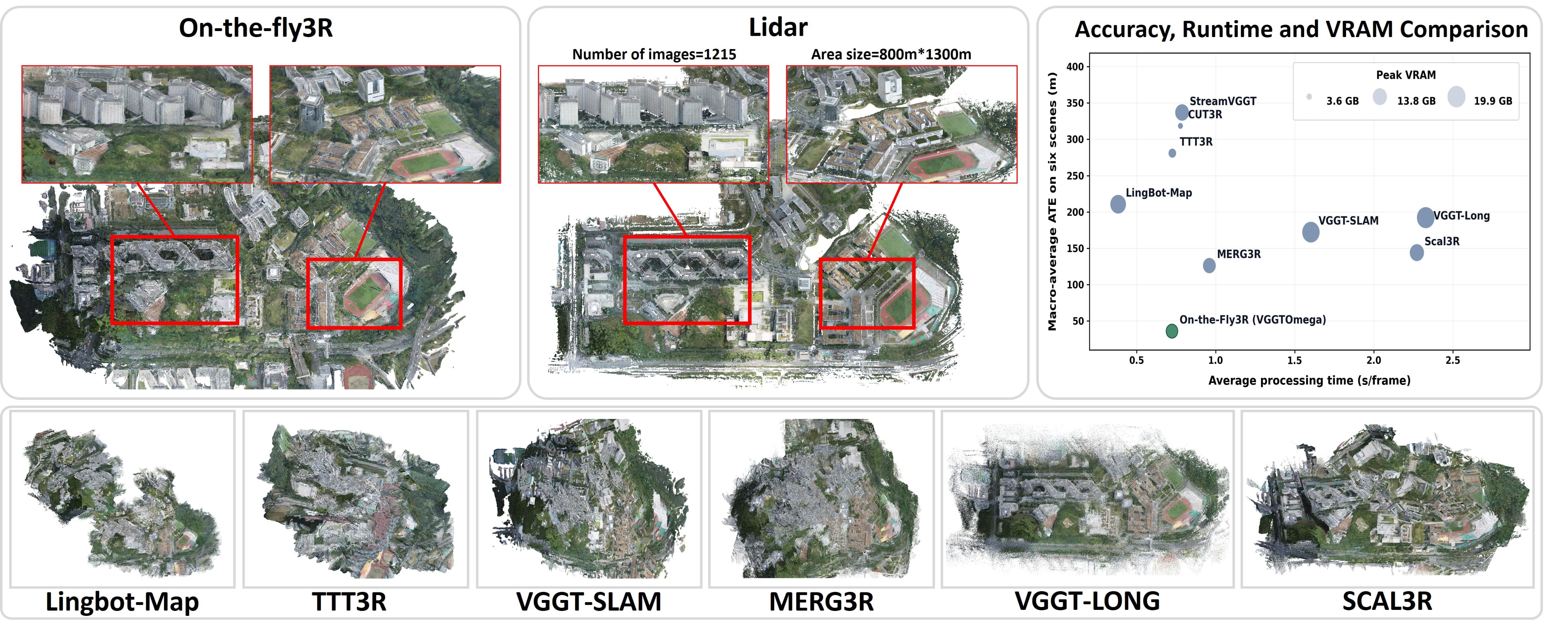}
    \vspace{-20pt}
    \captionof{figure}{Visual reconstruction results on the GauU-Scenes SZIT scene.}
    \label{fig:SZIT_overview}
    \vspace{-15pt}
\end{strip}

\begin{abstract}
While feed-forward 3D reconstruction (3R) offers efficient end-to-end modeling, its application in large-scale UAV mapping is hindered by the prohibitive memory cost of Transformer attention. Current scalable streaming 3R methods assume temporally and spatially continuous inputs, rendering them ineffective for the weakly ordered or unordered image streams common in cross-strip UAV operations. To address this, we propose \textit{On-the-Fly3R}, a training-free, progressive online 3D reconstruction framework for large-scale UAV images that upgrades various 3R backbones for large-scale UAV scenarios. Our method enables reconstruction from unordered inputs via retrieval-guided dynamic subset construction, which adaptively selects spatially relevant images. To further improve the robustness, a validation–rejection–retry mechanism is designed to guarantee global consistency, performing a pre-integration consistency check and automatically rejecting misaligned images and retrying with alternative subset. Finally, inspired by VSLAM, pose graph optimization based on the retrieval loop closure is employed to mitigate camera drift. Evaluations on several UAV benchmarks show that our \textit{On-the-Fly3R} successfully scales various 3R models to over 5,000 images across square-kilometer UAV scenes, delivering substantially superior accuracy compared to several SOTA streaming 3R methods. Code is available at \url{https://github.com/Sh1nZzz/On_the_Fly3R}

\end{abstract}

\section{INTRODUCTION} 
Recently, feed-forward 3D reconstruction (3R) methods, also recognized as 3D visual foundation models, have emerged as a powerful paradigm in the field of computer vision and photogrammetry. Representative methods such as DUSt3R~\cite{wang2024dust3r}, MASt3R~\cite{leroy2024mast3r}, and VGGT~\cite{wang2025vggt} can directly predict camera poses and dense 3D point clouds from uncalibrated images, offering a robust and efficient end-to-end pipeline from 2D observations to 3D scenes. However, scaling these methods to large-scale UAV scenarios remains a significant bottleneck. Built upon Transformer architectures, their global attention mechanisms cause computational and memory costs to scale quadratically with the number of input images. Consequently, on consumer-grade GPUs (e.g., a RTX 4090 with 24 GB VRAM), 3R methods frequently encounter out-of-memory failures when processing more than a few hundred images, severely limiting their practical deployment.

To improve the scalability of 3R methods, recent works generally fall into two categories: chunk-based and streaming reconstruction. Chunk-based methods~\cite{deng2025vggtlong, maggio2025vggtslam, zhang2026talo, xie2026scal3r} partition long image sequences into local chunks and align them via overlapping frames. Conversely, streaming-based approaches~\cite{wang2025spann3r, wang2025cut3r, zhuo2025streamvggt, chen2026ttt3r, yuan2026infinitevggt, chen2026lingbotmap} manage global context to progressively update the scene representation. While effective for continuous sequences, these methods inherently rely on strict assumptions of input with strong spatiotemporal continuity and local overlap. In large-scale UAV mapping tasks, factors such as cross-strip observations, multi-camera acquisition, and multi-sortie data collection inevitably result in unordered or weakly ordered image streams. Under such realistic conditions, existing scalable 3R methods degrade significantly, as they cannot effectively handle the lack of sequential continuity. A comparison with SOTA scalable 3R methods on a UAV dataset is shown in Fig. \ref{fig:SZIT_overview}.

Furthermore, current chunk-based methods typically accept the local chunk reconstructions produced by the feed-forward 3R model without explicit reliability verification. Under unordered or weakly ordered image conditions, a local chunk may contain images lacking effective co-visibility with the dominant scene. Such inconsistent observations can lead the model to hallucinate erroneous camera poses, scales, and 3D structures. Blindly integrating these anomalous local results into the global model, and subsequently using them as references for chunk alignment, triggers cascading error propagation. This not only causes incorrect registration and accumulated drift in subsequent chunks but ultimately catastrophically degrades the geometric consistency and robustness of the entire reconstruction. 

How can we effectively leverage the capabilities of existing 3R models under limited computational resources and extend them to large-scale reconstruction from unordered or weakly ordered UAV images? We argue that the crux lies in two aspects: constructing geometrically consistent local contexts for the VFMs, and promptly detecting and correcting local failures during online progressive reconstruction.

Driven by this insight, we propose \textit{On-the-Fly3R}, a training-free, progressive reconstruction framework that is compatible with diverse 3R models. Specifically, our method is first initialized with a small seed of images. Then, for each newly fly-in image, the co-visible images are retrieved from the reconstructed global map to dynamically construct a local subset, enabling the 3R models to perform feed-forward inference within a geometrically relevant local context. Each local prediction is then aligned to the global model via a confidence-weighted Sim(3) transformation. Crucially, we perform pose- and scale-based consistency checks before merging the local result into the global model. If the check fails, the framework automatically discards unreliable references and retries inference with a newly formed subset. Only newly added images and geometric results that pass validation are committed to the global map. Via this progressive pipeline, the proposed \textit{On-the-Fly3R} can process unordered or weakly ordered UAV images under a bounded memory footprint while effectively suppressing error propagation.

Our main contributions are summarized as follows:
 \begin{itemize} 
 \item We propose \textit{On-the-Fly3R}, a training-free, progressive feed-forward reconstruction framework, which can scale diverse 3R models to large scale UAV scenarios.
 \item A retrieval-guided dynamic subset construction strategy is introduced, that adaptively selects spatially relevant images to form geometrically consistent local inputs, effectively bridging the gap between feed-forward 3R models and unordered or weakly ordered images.
 \item A validation–rejection–retry mechanism is designed, explicitly verifying local reconstruction reliability prior to global integration. By automatically filtering out unreliable reference images upon consistency failure, this mechanism significantly enhances robustness and suppresses cascading error propagation.
 \end{itemize} 

\section{RELATED WORKS}
This section briefly reviews some relevant works.
\subsection{Traditional 3D Reconstruction}
Reconstructing 3D scenes from 2D images is a foundational problem in computer vision and photogrammetry. Traditional SfM systems, such as COLMAP~\cite{schonberger2016colmap}, recover camera poses and 3D points via incremental pipelines involving feature matching, geometric verification, and bundle adjustment, etc. These methods offer good scalability and geometric interpretability for large-scale image collections. Similarly, SLAM systems focus on real-time tracking and pose-graph optimization, with recent works like On-the-Fly series~\cite{zhan2025ontheflysfm,zhan2026ontheflysfm,ATDOM} pushing incremental reconstruction toward near-real-time performance via online retrieval.

Despite their maturity, traditional methods heavily rely on feature matching and computationally intensive iterative optimization, making them vulnerable in weakly textured or wide-baseline scenarios. Our framework draws inspiration from the retrieval-based view selection and geometric verification inherent in traditional incremental pipelines. However, instead of relying on fragile feature matching and slow optimization, we replace them with a frozen 3R model as a robust local reconstructor, enabling a more flexible and efficient progressive feed-forward paradigm.

\subsection{Feed-forward 3D Visual Foundation Models}
Recent feed-forward 3R models have introduced a revolutionary end-to-end paradigm for 3D reconstruction. DUSt3R~\cite{wang2024dust3r} pioneered the formulation of image-pair reconstruction as pointmap regression, which was subsequently enhanced by MASt3R~\cite{leroy2024mast3r} through learned geometric matching. To handle more general multi-view inputs, VGGT~\cite{wang2025vggt} and Fast3R~\cite{yang2025fast3r} extended this paradigm to improve multi-view inference efficiency. Further advancements include Pi3~\cite{wang2026pi3}, which bolsters robustness against input ordering and reference-view bias, and MapAnything~\cite{keetha2026mapanything}, which advances unified foundation models for general metric 3D reconstruction. While these models exhibit strong geometric priors and powerful end-to-end prediction capabilities, their reliance on global Transformer attention imposes severe memory bottlenecks, as computational and memory costs scale quadratically with the number of input images. Consequently, they struggle to scale directly to the thousands of images required for large-scale UAV mapping. Rather than retraining or modifying the backbone architecture, this paper treats the 3R model as a frozen local reconstructor, significantly enhancing its large-scale applicability through an external progressive framework.

\subsection{Scalable Feed-forward 3D Reconstruction}
To overcome the scalability limits of 3R models, recent works generally fall into two categories: streaming-based and chunk-based methods. Streaming methods process long sequences by maintaining compact states or contextual memories. Examples include the spatial memory in Spann3R~\cite{wang2025spann3r}, recurrent state in CUT3R~\cite{wang2025cut3r}, causal attention in StreamVGGT~\cite{zhuo2025streamvggt} and Stream3R~\cite{lan2026stream3r}, test-time training in TTT3R~\cite{chen2026ttt3r}, rolling memory in InfiniteVGGT~\cite{yuan2026infinitevggt}, and SLAM-inspired geometric context in LingBot-Map~\cite{chen2026lingbotmap}. However, the context construction in these methods inherently hinges on temporal order or sequential continuity, rendering them ineffective for unordered UAV image streams.

Alternatively, chunk-based methods adopt a divide-and-conquer strategy, partitioning sequences into local chunks followed by cross-chunk alignment. VGGT-Long~\cite{deng2025vggtlong} extends reconstruction to long sequences via overlapping alignment, while VGGT-SLAM~\cite{maggio2025vggtslam} integrates VGGT predictions with a SLAM backend using SL(4) optimization. To address spatial inconsistencies, TALO~\cite{zhang2026talo} employs Thin Plate Spline alignment, and Scal3R~\cite{xie2026scal3r} enhances long-range consistency via test-time global context memory. Notably, MERG3R~\cite{cheng2026merg3r} follows a divide-and-conquer strategy to handle large-scale unordered image collections and is one of the few related works that explicitly consider unordered inputs. More critically, existing chunk-based methods typically trust local reconstructions without explicit reliability verification. Under unordered conditions, local chunks may contain inconsistent observations, leading to erroneous poses and structures. Blindly integrating these anomalies triggers cascading error propagation, catastrophically degrading the global reconstruction.

In contrast, our \textit{On-the-Fly3R} is a training-free progressive framework that fundamentally departs from fixed chunks or temporal dependencies. It dynamically constructs geometrically relevant local subsets through retrieval, natively supporting unordered image collections. Moreover, to prevent the cascading failures observed in existing pipelines, we introduce a strict validation--rejection--retry mechanism. By explicitly verifying local consistency before global integration and automatically discarding unreliable references upon failure, our framework ensures robust reconstruction and effectively suppresses error propagation.

\begin{figure*}[t]
\vspace{-15pt}
\centering
\includegraphics[width=1.0\textwidth]{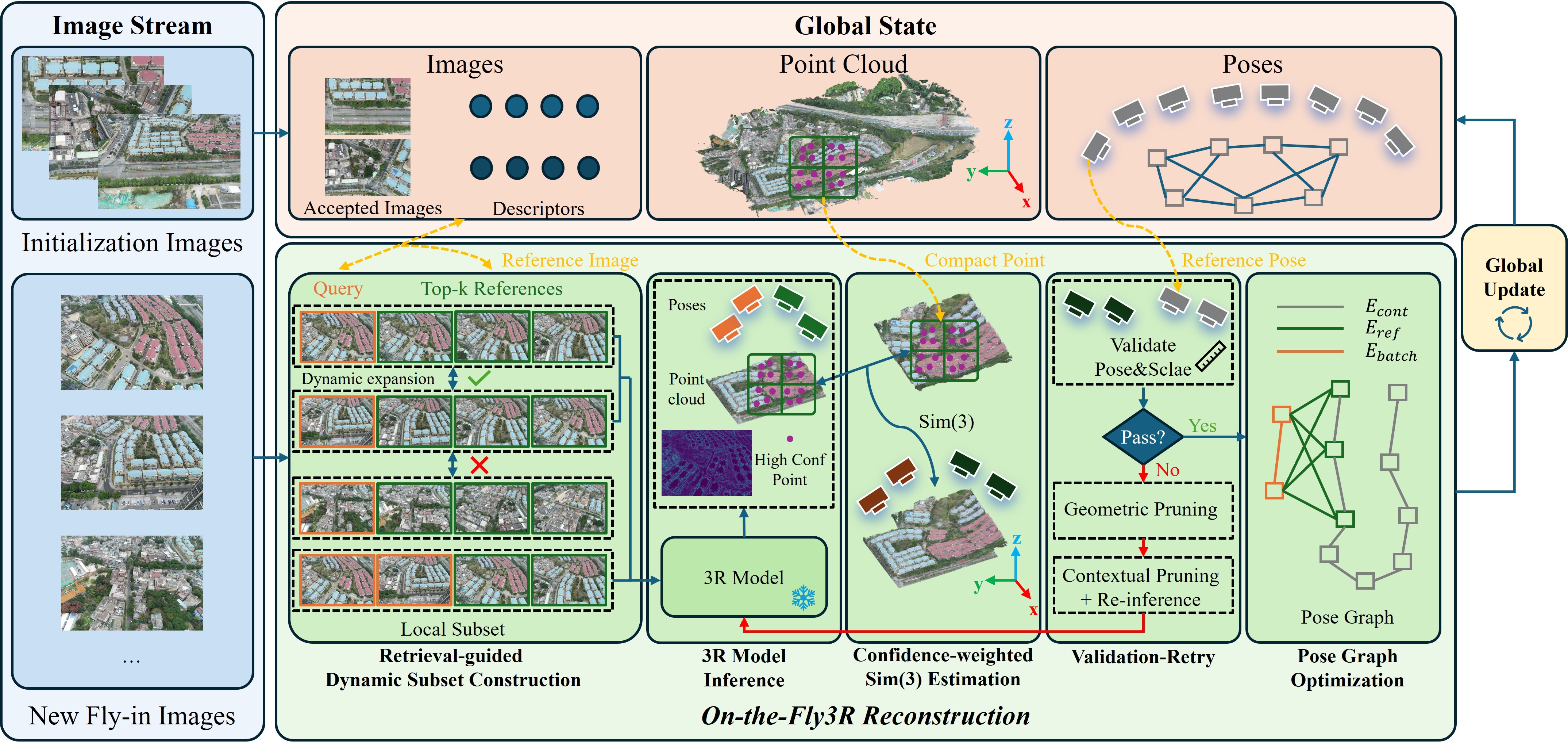}
\caption{Overview of the proposed \textit{On-the-Fly3R} framework.}
\label{fig:framework}
\vspace{-15pt}
\end{figure*}

\section{METHODOLOGY}
\label{sec:method}

\subsection{Overview}
\label{sec:method_overview}

Given an image set $\mathcal I=\{I_i\}_{i=1}^{N}$, where $I_i\in\mathbb R^{H_i\times W_i\times3}$, images arrive sequentially in an online fashion according to the capturing order. Our \textit{On-the-Fly3R} aims to incrementally estimate camera poses and reconstruct a dense point cloud upon the arrival of each new image, while explicitly preventing unreliable local reconstructions from corrupting the established global map.

The pipeline operates progressively, as Fig.~\ref{fig:framework} shows. It first initializes the global coordinate system and state using a seed set of images (Sec.~\ref{sec:initialization}). As new images fly in, a geometrically relevant local subset is dynamically constructed through image retrieval and fed into 3R model performs inference to obtain a local reconstruction (Sec.~\ref{sec:dynamic_subset}). A confidence-weighted $\mathrm{Sim}(3)$ transformation is then estimated to register the local reconstruction into the global map (Sec.~\ref{sec:alignment}). Crucially, a validation-retry mechanism verifies the registration consistency before merging; only validated results are incorporated into the global state (Sec.~\ref{sec:validation_retry}). Finally, inspired by VSLAM, a pose graph optimization is applied to mitigate accumulated drift (Sec.~\ref{sec:pgo}).

\subsection{Initialization}
\label{sec:initialization}

Similar to~\cite{ATDOM}, we initialize the framework with the first $N_0$ retrieval-connected images, following a retrieval procedure analogous to the query-subset construction described in (Sec.~\ref{sec:dynamic_subset}). These images form the initialization subset, denoted as  $\mathcal I_0=\{I_i\}_{i=1}^{N_0}$, and define $\mathcal A_0=\{1,\ldots,N_0\}$ as the initial accepted-frame set. To be compatible with various frozen 3R models $f$, we design a lightweight adapter $\Phi$ to unify its predictions into a standard representation:
\begin{equation}
    \Phi\!\left(f(\mathcal I_0)\right)
    =
    \left\{
    \widetilde{\mathbf T}_i,
    \mathbf K_i,
    \widetilde{\mathbf X}_i,
    \widetilde{\mathbf C}_i
    \right\}_{i\in\mathcal A_0}.
    \label{eq:vfm_output}
\end{equation}
where $\widetilde{\mathbf T}_i\in\mathbb R^{4\times4}$ is the camera-to-world transformation, $\mathbf K_i\in\mathbb R^{3\times3}$ is the camera intrinsic matrix, and $\widetilde{\mathbf X}_i\in\mathbb R^{H_i\times W_i\times3}$ and $\widetilde{\mathbf C}_i\in\mathbb R^{H_i\times W_i}$ are the per-pixel dense 3D point map and its confidence map, respectively. The coordinate system of this initial prediction serves as the global frame, from which we extract the initial global point cloud $\mathcal{P}_0$ using high-confidence points.

To maintain a bounded memory footprint, retaining full dense point maps for all historical frames is infeasible. Instead, we construct a compact geometric reference $\mathcal{H}_i$ for each accepted frame:
\begin{equation}
    \mathcal H_i
    =
    \left\{
    \left(\mathbf u_{in},\mathbf x^{g}_{in},c^{g}_{in}\right)
    \right\}_{n=1}^{P_i}.
    \label{eq:compact_reference}
\end{equation}
where $\mathbf u_{in}$ is a pixel location,$\mathbf x^{g}_{in}\in\mathbb R^3$ is the corresponding global 3D point, and $c^{g}_{in}$ is its confidence. Specifically, we partition each image into a regular grid and retain a fixed number of high-confidence points per cell, thereby controlling memory consumption while balancing geometric quality and spatial coverage. This compact representation later establishes geometric links between local reconstructions and the historical map.

Additionally, a frozen SupScene~\cite{shi2026supscene} encoder extracts an $L_2$-normalized descriptor $\mathbf d_i\in\mathbb R^D$ for retrieval. The per-frame state and the initialized global reconstruction state are formulated as
\begin{equation}
\begin{aligned}
    \mathcal F_i
    &=\left(\mathbf T_i^g,\mathbf K_i,\mathbf d_i,\mathcal H_i\right),\\
    \mathcal M_0
    &=\left(\{\mathcal F_i\}_{i\in\mathcal A_0},\mathcal P_0\right).
\end{aligned}
\label{eq:initial_state}
\end{equation}
Finally, we estimate the characteristic scene spacing $\ell_{\mathrm{scene}}$ from the median 3D distance between adjacent high-confidence points. This metric captures the local geometric scale of the scene and is later used to derive a scale-adaptive translation threshold for validation.

\subsection{Retrieval-Guided Dynamic Subset Construction}
\label{sec:dynamic_subset}

Since the online arrival order does not guarantee spatial continuity, a fixed sliding window may feed spatially disjoint images into local subset for 3R inference. To overcome this, we leverage the historical descriptors to dynamically select shared context for newly fly-in images. For a candidate query frame $q$, the SupScene~\cite{shi2026supscene} encoder produces a corresponding descriptor $\mathbf d_q$. Its cosine similarity to an accepted historical frame $r$ is
\begin{equation}
    s(q,r)=\mathbf d_q^{\top}\mathbf d_r.
    \label{eq:retrieval_similarity}
\end{equation}
Given the current global state $\mathcal{M}_{k-1}$, the top-$k_r$ historical reference set for $q$ is:
\begin{equation}
    \mathcal N_{k_r}(q)
    =
    \operatorname*{TopK}_{r\in\mathcal A_{k-1}}^{k_r}
    s(q,r),
    \label{eq:retrieval}
\end{equation}
To process images in batches, the dynamic query subset $\mathcal{Q}$ expands sequentially. For a candidate frame $q$, we evaluate its compatibility with the current query set $\mathcal{Q}$ using two metrics: the reference-overlap count $o(q,\mathcal{Q})$ and the maximum visual similarity $s_{\max}(q,\mathcal{Q})$ to existing queries:
\begin{equation}
\begin{aligned}
    o(q,\mathcal Q)
    &=\left|\mathcal N_{k_r}(q)\cap\mathcal D(\mathcal Q)\right|,\\
    s_{\max}(q,\mathcal Q)
    &=\max_{q'\in\mathcal Q}\mathbf d_q^{\top}\mathbf d_{q'}.
\end{aligned}
\label{eq:batch_compatibility}
\end{equation}
where $\mathcal{D}(\mathcal{Q})$ is the dominant reference set aggregated from $\mathcal{Q}$. A candidate is admitted to $\mathcal{Q}$ if and only if:
\begin{equation}
    o(q,\mathcal Q)\geq\tau_o
    \quad\land\quad
    s_{\max}(q,\mathcal Q)\geq\tau_s.
    \label{eq:batch_admission}
\end{equation}
This condition ensures that a newly arrived image is both visually coherent with the current batch and shares sufficient geometric support with it. Expansion halts when the next image is incompatible or when the batch reaches its maximum capacity $B_{\max}$, yielding the final query set $\mathcal{Q}_k$.

We then aggregate the retrieval results of all frames in $\mathcal{Q}_k$, prioritizing historical frames supported by multiple queries with high similarity, to form the reference subset $\mathcal{R}_k$. The final local subset fed to the 3R model is:
\begin{equation}
    \mathcal B_k=\mathcal R_k\cup\mathcal Q_k.
    \label{eq:vfm_context}
\end{equation}
Applying the same adapter $\Phi$ as in initialization, we obtain the unified local reconstruction:
\begin{equation}
    \Phi\!\left(f(\mathcal B_k)\right)
    =
    \left\{
    \widetilde{\mathbf T}_i^{(k)},
    \mathbf K_i,
    \widetilde{\mathbf X}_i^{(k)},
    \widetilde{\mathbf C}_i^{(k)}
    \right\}_{i\in\mathcal B_k}.
    \label{eq:local_vfm_output}
\end{equation}

\subsection{Confidence-Weighted $\mathrm{Sim}(3)$ Estimation}
\label{sec:alignment}

The reference frames $\mathcal R_k$ occur in both the current local reconstruction and the historical global state, serving as  geometric links between the two coordinate systems. For the compact geometric reference $\mathcal H_r$ of a reference frame $r$, we utilize its stored pixel locations $\mathbf u_{rn}$ to sample the current local 3D points $\widetilde{\mathbf x}_{rn}^{(k)}$ from the point map, pairing them with the historical global points $\mathbf x_{rn}^{g}$. The correspondence weight is defined as the geometric mean of their confidences:
\begin{equation}
    w_{rn}^{(k)}
    =
    \sqrt{\widetilde c_{rn}^{(k)}c_{rn}^{g}}.
    \label{eq:correspondence_weight}
\end{equation}

After collecting all valid correspondences $\mathcal Z_k$ across all reference frames $\mathcal{R}_k$, we estimate the similarity transformation $\mathcal S_k=(\sigma_k,\mathbf R_k,\mathbf t_k)\in\mathrm{Sim}(3)$ to register the $k$-th local reconstruction to the global map:
\begin{equation}
\begin{aligned}
    \min_{\substack{
        \sigma_k>0,\ \mathbf R_k\in\mathrm{SO}(3),\\
        \mathbf t_k\in\mathbb R^3}}
    \sum_{(r,n)\in\mathcal Z_k}
    w_{rn}^{(k)}\,
    \rho_{\kappa}\!\left(
    \left\|
    \sigma_k\mathbf R_k\widetilde{\mathbf x}_{rn}^{(k)}
    +\mathbf t_k-\mathbf x_{rn}^{g}
    \right\|_2
    \right),
\end{aligned}
\label{eq:sim3}
\end{equation}
where $\rho_{\kappa}$ is a Huber robust loss. We solve this via a two-stage pipeline: initializing with a confidence-weighted closed-form solution, followed by Iteratively Reweighted Least Squares (IRLS) to suppress outlier correspondences.

\subsection{Validation-Retry Mechanism and Global Update}
\label{sec:validation_retry}

The reliability of local registration is inherently vulnerable to two factors. On the one hand, 3R models' predictions for the same reference image are inconsistent under different contexts. On the other hand, appearance-based retrieval may introduce historical references that are visually similar but spatially unrelated. Both can yield a numerically estimable but geometrically spurious $\mathrm{Sim}(3)$. Therefore, to prevent error propagation, before updating the global state, we enforce a strict validation-retry protocol before global integration.

Specifically, the local camera poses of the reference frames $\mathcal{R}_k$ are transformed into the global frame via the estimated $\mathrm{Sim}(3)$ and compared against the camera poses stored in the global state. For each reference frame $r$, the translation and rotation residuals are: 
\begin{equation}
\begin{aligned}
    e_r^t
    &=\left\|\widehat{\mathbf t}_r^g-\mathbf t_r^g\right\|_2,\\
    e_r^R
    &=\arccos\!\left(
    \frac{
    \operatorname{tr}\!\left(
    \widehat{\mathbf R}_r^g(\mathbf R_r^g)^{\top}
    \right)-1}{2}
    \right).
\end{aligned}
\label{eq:pose_validation_residuals}
\end{equation}
The system jointly evaluates these residuals and verifies that the estimated scale $\sigma_k$ is physically plausible. The translation threshold is adaptively scaled by the characteristic scene spacing estimated during initialization:
\begin{equation}
    \tau_t=\alpha\ell_{\mathrm{scene}}.
    \label{eq:adaptive_translation_threshold}
\end{equation}

If the initial validation fails, the system triggers a two-stage retry mechanism:
\begin{itemize}
    \item \textbf{Geometric Pruning:} We estimate an independent $\mathrm{Sim}(3)$ for each reference frame and analyze their consistency to identify and remove anomalous historical references. The system then re-estimates and validates $\mathcal{S}_k$ using the remaining references \textit{without} re-running the 3R model inference.
    \item \textbf{Contextual Pruning:} Inspired by the observation in~\cite{wang2026vggtomega} that register tokens can capture global information, we use them when geometric pruning fails. We compare the 3R model register tokens of the reference and query frames to identify semantically incompatible references. These are discarded, and the 3R model inference, registration, and validation are \textit{repeated once} with the pruned subset.
\end{itemize}
Only local results passing either the initial check or a retry stage are committed to the global state. Let $\Delta\mathcal P_k$ denote the points fused from the new query frames in the $k$-th update. The global state updates as:
\begin{equation}
\begin{aligned}
    \mathcal A_k
    &=\mathcal A_{k-1}\cup\mathcal Q_k,\\
    \mathcal P_k
    &=\mathcal P_{k-1}\cup\Delta\mathcal P_k,\\
    \mathcal M_k
    &=\left(\{\mathcal F_i\}_{i\in\mathcal A_k},\mathcal P_k\right).
\end{aligned}
\label{eq:state_update}
\end{equation}
Each newly accepted frame state $\mathcal F_i$ contains its global camera pose, intrinsics, retrieval descriptor, and compact geometric reference. If both retry stages fail, the entire local subset is rejected and $\mathcal M_k=\mathcal M_{k-1}$, effectively isolating the unreliable reconstruction and preventing cascading errors.

\subsection{Pose Graph Optimization}
\label{sec:pgo}

While the validation mechanism ensures local reliability, successive local-to-global registrations may still accumulate drift. To enforce global consistency, we construct a pose graph $\mathcal G=(\mathcal V,\mathcal E)$ for pose refinement. The node set $\mathcal V=\{v_i\mid i\in\mathcal A_K\}$ corresponds to all accepted frames, and each node is associated with an optimizable camera pose $\mathbf T_i$. The edge set consists of three types of constraints:
\begin{equation}
    \mathcal E
    =
    \mathcal E_{\mathrm{ref}}
    \cup\mathcal E_{\mathrm{batch}}
    \cup\mathcal E_{\mathrm{cont}}.
    \label{eq:pgo_edges}
\end{equation}
Here, $\mathcal E_{\mathrm{ref}}$ connects historical reference frames to current query frames (acting as loop closures for large temporal gaps); $\mathcal E_{\mathrm{batch}}$ connects query frames within the same local 3R reconstruction; and $\mathcal E_{\mathrm{cont}}$ connects temporally adjacent arriving images that satisfy retrieval-association and motion-plausibility checks. Edge weights in $\mathbf{\Omega}_{ij}$ are dynamically assigned based on alignment robustness (inlier ratio and residual magnitude).

The final camera poses are optimized on $\mathrm{SE}(3)$ via robust least squares:
\begin{equation}
\begin{aligned}
    \min_{\{\mathbf T_i\}}
    &\sum_{(i,j)\in\mathcal E}
    \rho_H\!\left(
    \mathbf r_{ij}^{\top}\mathbf\Omega_{ij}\mathbf r_{ij}
    \right),\\
    \mathbf r_{ij}
    &=\operatorname{Log}\!\left(
    \overline{\mathbf T}_{ij}^{-1}
    \mathbf T_i^{-1}\mathbf T_j
    \right).
\end{aligned}
\label{eq:pgo}
\end{equation}
where $\overline{\mathbf T}_{ij}$ is the relative-pose observation on an edge, $\mathbf\Omega_{ij}$ is the edge information matrix encoding constraint reliability, and $\rho_H$ is a Huber robust kernel.

\section{EXPERIMENT}

\subsection{Datasets and Evaluation Metrics}
\label{sec:datasets_metrics}

We evaluate \textit{On-the-Fly3R} on two large-scale UAV benchmarks and one popular indoor dataset to assess both domain-specific performance and cross-domain generalization. GauU-SceneV2~\cite{xiong2024gauuscenev2} features four weakly ordered scenes ranging from 424 to 1,500 images, providing high-quality LiDAR ground truth and reference camera poses. UrbanScene~\cite{lin2022urbanscene3d} comprises two massive scenes, Residence (2,582 images) and Campus (5,871 images), with dense reference point clouds generated by the commercial software - iTwin. Collectively, these datasets present significant real-world challenges, including multi-strip flight patterns, varying viewing angles, repetitive structures, and substantial scale variations. Additionally, we include the 7Scenes~\cite{shotton2013scene} to examine our generalization capability of indoor environments.

For camera pose evaluation, we report four standard root-mean-square error (RMSE) metrics: 
Absolute Translation Error (ATE, in meters) and Absolute Rotation Error (ARE, in degrees) measure global discrepancies; 
Relative Translation Error (RTE) quantifies the angular deviation between the predicted and ground-truth translation vectors of consecutive frames; 
and Relative Rotation Error (RRE, in degrees) measures their relative rotational discrepancy. For dense reconstruction, we report accuracy (Acc.), completeness (Comp.), and Chamfer distance (CD) against the referenced point clouds.

\subsection{Implementation details}
\label{sec:Implementation details}

All experiments are executed on a machine  equipped with dual Intel Xeon Gold 6133 CPUs and a single NVIDIA RTX 4090 GPU (24~GB VRAM) running Ubuntu 22.04. Consistent with our training-free design, all 3R models are kept strictly frozen without any fine-tuning; \textbf{Pi3x}~\cite{wang2026pi3} serves as the default backbone unless otherwise specified. Regarding the settings of hyperparameters, we initialize the global map with $N_0=30$ images. During progressive reconstruction, each dynamic batch admits up to $B_{\max}=5$ new query images, retrieves $k_r=5$ historical references, and requires a minimum of 3 valid references to ensure robust local-to-global alignment. To strictly bound memory consumption, we retain up to 5,000 compact geometric reference points per accepted frame. Final camera poses are globally refined using a pose-graph optimization backend implemented in GTSAM.

To ensure a strictly fair comparison, all baseline methods share identical input images, ground-truth references, and evaluation protocols. Runtime is normalized per input image, and peak GPU memory is recorded over the entire reconstruction pipeline.

\subsection{Comparison with other SOTA 3R Methods}

We compare \textit{On-the-Fly3R} against eleven state-of-the-art scalable 3R methods, encompassing both streaming-based (LingBot-Map~\cite{chen2026lingbotmap}, StreamVGGT~\cite{zhuo2025streamvggt}, CUT3R~\cite{wang2025cut3r}, TTT3R~\cite{chen2026ttt3r}, InfiniteVGGT~\cite{yuan2026infinitevggt}) and chunk-based (Scal3R~\cite{xie2026scal3r}, VGGT-SLAM~\cite{maggio2025vggtslam}, VGGT-Long~\cite{deng2025vggtlong}, TALO~\cite{zhang2026talo}, FastVGGT~\cite{shen2025fastvggt}, MERG3R~\cite{cheng2026merg3r}) paradigms. All methods are fed the identical complete image sequences. Since our validation mechanism may reject unreliable local updates to prevent cascading errors, we report both the \textit{returned-pose} metrics (evaluating only the poses the method successfully outputs) and the \textit{coverage rate}. To ensure a strictly fair comparison free from selective output bias, we further evaluate all methods on a \textit{common frame set} consisting of the 11,595 images (95.39\% of the total 12,155 inputs) successfully accepted by \textit{On-the-Fly3R} (Pi3x) across the six outdoor UAV scenes.

Table~\ref{tab:pose_per_scene} reports the per-scene ATE and the overall pose coverage. Methods like FastVGGT and InfiniteVGGT suffer from Out-Of-Memory (OOM) errors on larger scenes, highlighting the severe memory bottleneck of native global inference. TALO fails to complete most scenes (21.24\% coverage). In contrast, all three variants of \textit{On-the-Fly3R} maintain high coverage (89\%--96\%) while achieving significantly lower errors. Notably, \textit{On-the-Fly3R} (Pi3x) achieves the most stable and comprehensive performance across all scenes, while the VGGT-Omega variant attains the highest coverage and excels in specific large-scale scenes like Campus.

\begin{table}[t]
\centering
\caption{Per-scene ATE (m) and pose coverage (\%) across six outdoor scenes. Coverage denotes the ratio of successfully reconstructed poses. Best and second-best errors are highlighted in bold and underlined, respectively. ''--'' indicates failure to reconstruct, and ''OOM'' denotes out-of-memory.}
\label{tab:pose_per_scene}
\setlength{\tabcolsep}{2.8pt}
\renewcommand{\arraystretch}{1.10}
\resizebox{\columnwidth}{!}{%
\begin{tabular}{lccccccc}
\hline
Method & \shortstack{HAV\\424} & \shortstack{SMBU\\563} &
\shortstack{SZIT\\1215} & \shortstack{SZTU\\1500} & \shortstack{Residence\\2582} &
\shortstack{Campus\\5871} & \shortstack{Coverage\\(\%)} \\
\hline
LingBot-Map & 180.43 & 135.55 & 295.06 & 176.75 & 73.04 & 404.43 & 100.00 \\
StreamVGGT & 263.77 & 269.30 & 384.35 & 470.77 & 134.32 & 499.33 & 100.00 \\
CUT3R & 210.17 & 260.63 & 365.24 & 461.85 & 133.19 & 480.23 & 100.00 \\
TTT3R & 123.59 & 207.09 & 313.12 & 468.26 & 115.58 & 458.93 & 100.00 \\
Scal3R & 51.21 & 40.77 & 168.44 & 211.59 & 44.76 & 349.00 & 100.00 \\
VGGT-SLAM & 115.28 & 103.24 & 87.25 & 241.94 & 16.49 & 470.09 & 100.00 \\
VGGT-Long & 54.16 & 185.47 & 166.04 & 367.10 & 44.38 & 336.06 & 100.00 \\
TALO & -- & -- & -- & -- & 103.92 & -- & 21.24 \\
FastVGGT & OOM & OOM & OOM & OOM & OOM & OOM & 0.00 \\
InfiniteVGGT & 216.15 & 167.83 & 310.41 & 382.03 & OOM & OOM & 30.46 \\
MERG3R & 50.02 & \textbf{9.80} & 145.13 & 55.33 & 5.74 & 492.95 & 100.00 \\
On-the-Fly3R (Pi3) & 7.73 & 17.22 & \underline{23.72} & \underline{47.18} & 3.28 & 42.29 & 89.81 \\
On-the-Fly3R (Pi3x) & \underline{4.20} & \underline{14.47} & 29.27 & \textbf{28.82} & \underline{1.65} & \underline{32.95} & 95.39 \\
\textbf{On-the-Fly3R (VGGT-Omega)} & \textbf{3.64} & 93.54 & \textbf{5.22} & 97.46 & \textbf{1.25} & \textbf{16.00} & 96.26 \\
\hline
\end{tabular}}
\vspace{-15pt}
\end{table}

\begin{table}[t]
\centering
\caption{Six-scene macro-average pose errors evaluated on the common frame set accepted by \textit{On-the-Fly3R} (Pi3x). Methods that failed to reconstruct the common set are omitted.}
\label{tab:pose_macro}
\setlength{\tabcolsep}{3.3pt}
\renewcommand{\arraystretch}{1.10}
\resizebox{\columnwidth}{!}{%
\begin{tabular}{lcccc}
\hline
Method & ATE (m)$\downarrow$ & ARE ($^\circ$)$\downarrow$ & RTE ($^\circ$)$\downarrow$ & RRE ($^\circ$)$\downarrow$ \\
\hline
LingBot-Map & 205.62 & 26.02 & 43.01 & 5.76 \\
StreamVGGT & 328.54 & 126.41 & 74.52 & 15.81 \\
CUT3R & 311.52 & 88.16 & 46.25 & 13.29 \\
TTT3R & 270.91 & 47.44 & 48.32 & 13.81 \\
Scal3R & 142.88 & 29.23 & 18.11 & 3.01 \\
VGGT-SLAM & 187.12 & 29.22 & 22.78 & 6.49 \\
VGGT-Long & 187.88 & 50.10 & 22.78 & 8.20 \\
MERG3R & 124.42 & 26.29 & 54.11 & 22.89 \\
\textbf{On-the-Fly3R (Pi3x)} & \textbf{18.56} & \textbf{2.78} & \textbf{10.85} & \textbf{1.52} \\
\hline
\end{tabular}}
\vspace{-10pt}
\end{table}

Table~\ref{tab:pose_macro} presents the macro-average results on the common frame set. By controlling for the evaluation mask, we eliminate the bias of selective reporting. \textit{On-the-Fly3R} (Pi3x) drastically reduces the ATE from 124.42~m (the strongest baseline, MERG3R) to 18.56~m, while simultaneously achieving superior ARE, RTE, and RRE. This confirms that our retrieval-guided context and validation mechanism significantly resolve the catastrophic drifts inherent in existing scalable methods.

To verify that our progressive framework is not strictly limited to large-scale outdoor UAV imagery, we evaluate its generalization capability on the indoor 7Scenes dataset. As shown in Table~\ref{tab:seven_scenes}, \textit{On-the-Fly3R} with VGGT-Omega achieves the lowest ATE across all seven scenes, reducing the scene-averaged ATE from 5.32~cm (MERG3R) to an impressive 2.30~cm. The Pi3 and Pi3x variants also remain highly competitive with the strongest baselines. This demonstrates the plug-and-play nature of our framework, which can seamlessly upgrade diverse 3R models for robust indoor reconstruction.

\begin{table}[t]
\centering
\caption{ATE (cm) on 7Scenes(sequence 01, temporal stride 5). Best and second-best results are highlighted in bold and underlined, respectively.}
\label{tab:seven_scenes}
\setlength{\tabcolsep}{4.0pt}
\renewcommand{\arraystretch}{1.10}
\resizebox{\columnwidth}{!}{%
\begin{tabular}{lcccccccc}
\hline
Method & Chess & Fire & Heads & Office & Pumpkin & Kitchen & Stairs & Avg. \\
\hline
LingBot-Map & 3.93 & 3.49 & 3.16 & 10.48 & 13.82 & 5.49 & 14.39 & 7.82 \\
StreamVGGT & 59.49 & 56.65 & 26.45 & 50.79 & 56.66 & 40.66 & 79.30 & 52.86 \\
CUT3R & 59.51 & 9.72 & 19.09 & 32.03 & 38.64 & 20.98 & 31.73 & 30.24 \\
TTT3R & 7.63 & 4.48 & 5.42 & 11.31 & 18.30 & 7.10 & 8.23 & 8.93 \\
Scal3R & 4.81 & 3.29 & \underline{1.79} & 10.40 & 14.09 & 5.48 & \underline{1.95} & 5.97 \\
VGGT-SLAM & 4.00 & 2.80 & 3.66 & 9.53 & 15.08 & 5.00 & 2.17 & 6.04 \\
VGGT-Long & 4.93 & 3.20 & 2.18 & 9.41 & 14.86 & 4.51 & 4.49 & 6.23 \\
TALO & 30.30 & 7.68 & 4.11 & 13.71 & 14.84 & 3.97 & 2.58 & 11.03 \\
FastVGGT & 4.09 & 3.18 & 1.91 & 11.24 & 15.43 & 5.55 & 3.40 & 6.40 \\
InfiniteVGGT & 7.40 & 3.50 & 4.56 & 14.75 & 14.08 & 8.61 & 30.09 & 11.86 \\
MERG3R & \underline{3.84} & \underline{2.63} & 1.85 & 8.29 & 13.94 & 3.91 & 2.76 & \underline{5.32} \\
On-the-Fly3R (Pi3) & 3.90 & 3.38 & 3.80 & 8.19 & \underline{13.42} & 3.38 & 2.95 & 5.57 \\
On-the-Fly3R (Pi3x) & 4.04 & 3.73 & 3.55 & \underline{8.11} & 14.23 & \underline{3.33} & 2.38 & 5.62 \\
\textbf{On-the-Fly3R (VGGT-Omega)} & \textbf{1.64} & \textbf{2.02} & \textbf{1.72} & \textbf{2.38} & \textbf{4.80} & \textbf{1.97} & \textbf{1.58} & \textbf{2.30} \\
\hline
\end{tabular}}
\vspace{-15pt}
\end{table}

\begin{figure}[h!]
\centering
\includegraphics[width=\columnwidth]{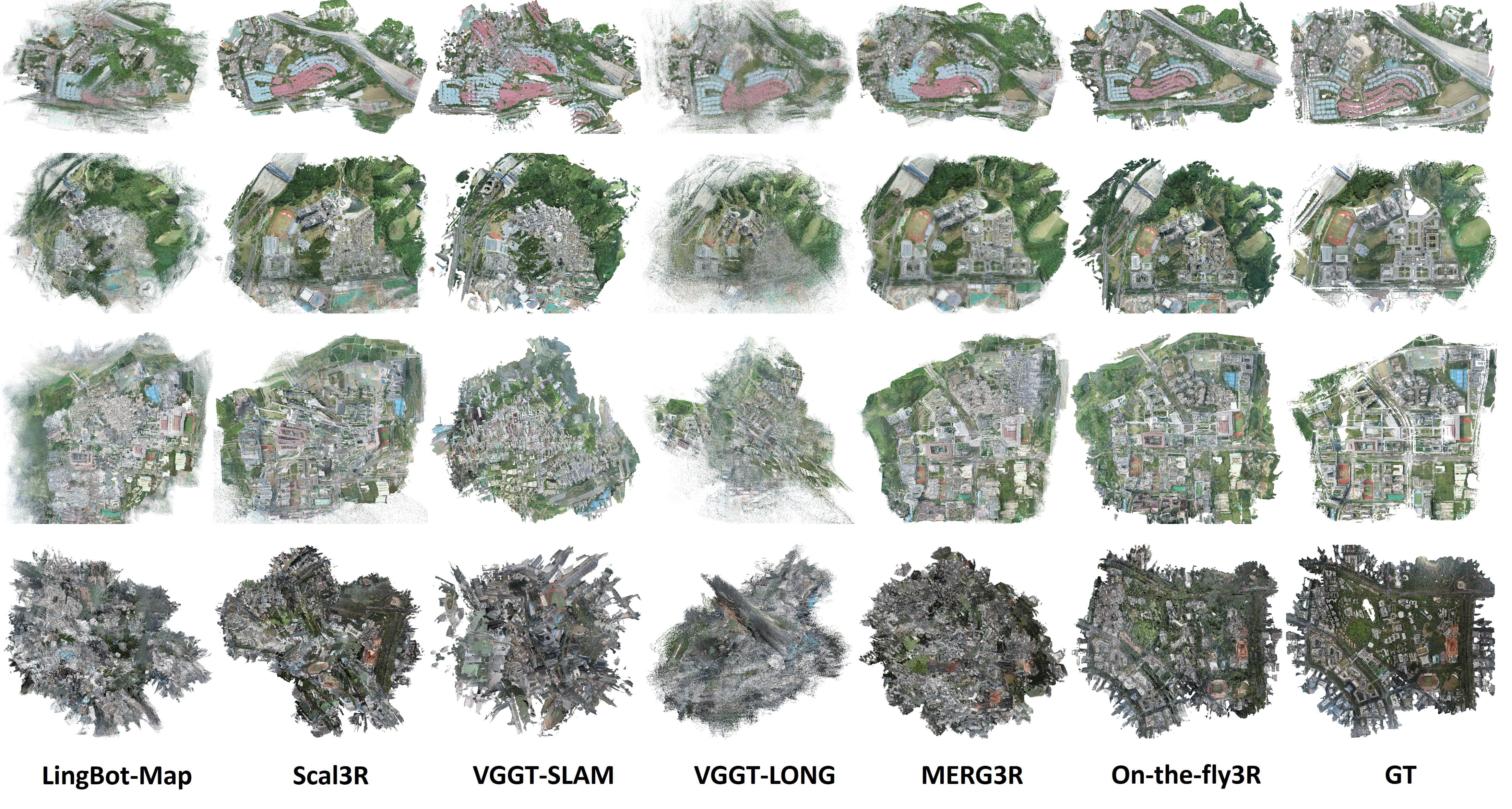}
\caption{Qualitative comparison of dense reconstruction.} 
\label{fig:compare_fig}
\vspace{-6pt}
\end{figure}

Table~\ref{tab:geometry_efficiency} evaluates the trade-off between dense reconstruction quality, runtime, and peak GPU memory. \textit{On-the-Fly3R} (Pi3x) achieves a mean Chamfer Distance (CD) of 3.85~m, which is substantially lower than the best complete baseline (MERG3R at 16.41~m) and the fastest baseline (LingBot-Map at 18.07~m). While LingBot-Map is faster (0.38~s/frame), its reconstruction accuracy is severely compromised. In contrast, our method maintains a highly competitive runtime (0.72--0.81~s/frame), comparable to streaming methods and significantly faster than chunk-based approaches, while strictly operating within the 24~GB GPU memory budget. A visual comparison result can be found in Fig.~\ref{fig:compare_fig}, showing our point cloud is the most consistent with the ground truth. This demonstrates a Pareto-optimal trade-off, delivering massive accuracy gains with only a marginal computational overhead.

\begin{table}[t]
\centering
\caption{Dense reconstruction quality, runtime, and peak-memory on the six outdoor UAV scenes. Point-cloud metrics and runtime are macro-averaged across completed scenes; memory is the maximum recorded peak. ''$\dagger$'' denotes an incomplete point-cloud result.}
\label{tab:geometry_efficiency}
\setlength{\tabcolsep}{2.2pt}
\renewcommand{\arraystretch}{1.10}
\resizebox{\columnwidth}{!}{%
\begin{tabular}{lcccccc}
\hline
Method & PC scenes & Acc.$\downarrow$ & Comp.$\downarrow$ & CD$\downarrow$ & \shortstack{Time\\(s/frame)}$\downarrow$ & \shortstack{Peak\\(GB)}$\downarrow$ \\
\hline
LingBot-Map  & 6/6 & 18.12 & 18.02 & 18.07 & \textbf{0.382} & 18.40 \\
StreamVGGT$^{\dagger}$ & 5/6 & 47.80 & 46.24 & 47.02 & 0.787 & 13.78 \\
CUT3R        & 6/6 & 37.78 & 57.84 & 47.81 & 0.776 & \textbf{3.58} \\
TTT3R        & 6/6 & 30.80 & 41.84 & 36.32 & 0.725 & \underline{6.20} \\
Scal3R       & 6/6 & 20.39 & 30.68 & 25.54 & 2.272 & 14.38 \\
VGGT-SLAM    & 6/6 & 17.07 & 9.81 & 13.44 & 1.602 & 21.46 \\
VGGT-Long    & 6/6 & 19.64 & 18.31 & 18.97 & 2.328 & 19.87 \\
MERG3R       & 6/6 & 24.69 & 8.12 & 16.41 & 0.959 & 16.36 \\
On-the-Fly3R (Pi3) & 6/6 & \underline{5.47} & 5.77 & 5.62 & 0.736 & 9.90 \\
\textbf{On-the-Fly3R (Pi3x)} & 6/6 & \textbf{4.47} & \underline{3.22} & \textbf{3.85} & 0.814 & 17.56 \\
On-the-Fly3R (VGGT-Omega) & 6/6 & 7.26 & \textbf{2.44} & \underline{4.85} & \underline{0.723} & 9.90 \\
\hline
\end{tabular}}
\vspace{-15pt}
\end{table}

\subsection{Comparison with Vanilla 3R Methods}
\label{sec:vfm_comparison}

To quantify the accuracy cost of scaling, we compare \textit{On-the-Fly3R} against the native joint inference of various 3R models. Under the 24~GB memory constraint, we restrict the input to 58 images to ensure all backbones can complete native inference without OOM. As shown in Table~\ref{tab:vfm_comparison}, replacing global joint inference with our progressive local updates introduces only a marginal accuracy gap. \textit{On-the-Fly3R} even outperforms native inference in 7 out of 10 model-scene combinations. This proves that our retrieval-guided context and validation mechanism successfully preserve the powerful geometric priors of the frozen VFMs while extending their capacity to massive scenes.

\begin{table}[t]
\centering
\caption{Comparison between vanilla 3R methods joint inference and \textit{On-the-Fly3R} using an identical 58-image input. Bold indicates cases where \textit{On-the-Fly3R} outperforms native inference.}
\label{tab:vfm_comparison}
\setlength{\tabcolsep}{2.3pt}
\renewcommand{\arraystretch}{1.10}
\resizebox{\columnwidth}{!}{%
\begin{tabular}{lccccc}
\hline
Backbone & Mode & \shortstack{SZIT\\ATE}$\downarrow$ & \shortstack{Residence\\ATE}$\downarrow$ & \shortstack{SZIT\\CD}$\downarrow$ & \shortstack{Residence\\CD}$\downarrow$ \\
\hline

\multirow{2}{*}{MapAnything}
& Vanilla & 5.73 & 3.62 & 1.90 & 1.82 \\
& On-the-Fly3R & \textbf{5.68} & \textbf{2.42} & 2.24 & 2.14 \\
\hline

\multirow{2}{*}{Pi3}
& Vanilla & 3.38 & 1.92 & 1.41 & 1.79 \\
& On-the-Fly3R & 7.81 & 2.30 & 2.50 & 2.02 \\
\hline

\multirow{2}{*}{Pi3x}
& Vanilla & 2.17 & 1.34 & 1.32 & 1.33 \\
& On-the-Fly3R & 3.17 & \textbf{1.20} & 1.63 & 1.35 \\
\hline

\multirow{2}{*}{VGGT}
& Vanilla & 5.88 & 8.46 & 1.81 & 2.60 \\
& On-the-Fly3R & \textbf{3.60} & \textbf{2.56} & \textbf{1.74} & \textbf{1.50} \\
\hline

\multirow{2}{*}{VGGT-Omega}
& Vanilla & 2.19 & 1.74 & 1.47 & 1.34 \\
& On-the-Fly3R & \textbf{2.11} & \textbf{1.13} & 1.61 & \textbf{1.21} \\
\hline

\end{tabular}}
\vspace{-10pt}
\end{table}

\subsection{Ablation Study}
\label{sec:ablation}

\begin{figure*}[h!]
\vspace{-15pt}
\centering
\includegraphics[width=\textwidth]{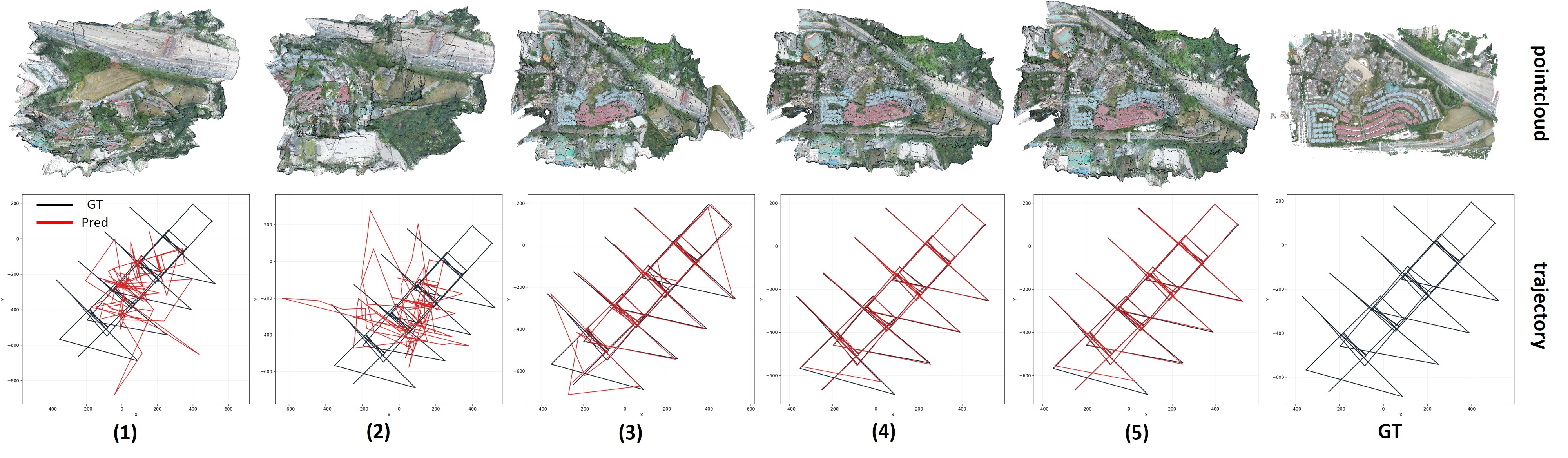}
\caption{Visual results of ablation studies. 
}
\label{fig:ablation_fig}
\vspace{-15pt}
\end{figure*}

We conduct ablations with VGGT-Omega as backbone 3R model on HAV and Residence to isolate the contributions of three core components: retrieval-guided dynamic batching (Dyn.), validation with reference-pruning retry (V\&R), and pose graph optimization (PGO). Table~\ref{tab:ablation} reports the four pose RMSE metrics under five cases, and a corresponding qualitative result is illustrated by Fig.~\ref{fig:ablation_fig}.

\begin{table}[t]
\centering
\caption{Ablation study with VGGT-Omega on HAV and Residence. Dyn.: Dynamic batching; V\&R: Validation \& Retry; PGO: Pose Graph Optimization.}
\label{tab:ablation}
\setlength{\tabcolsep}{4.0pt}
\renewcommand{\arraystretch}{1.10}
\resizebox{\columnwidth}{!}{%
\begin{tabular}{cccccccccccc}
\hline
& & & & \multicolumn{4}{c}{HAV} & \multicolumn{4}{c}{Residence} \\
Case & Dyn. & V\&R & PGO & ATE$\downarrow$ & ARE$\downarrow$ & RTE$\downarrow$ & RRE$\downarrow$ & ATE$\downarrow$ & ARE$\downarrow$ & RTE$\downarrow$ & RRE$\downarrow$ \\
\hline
(1) & \ding{56} & \ding{56} & \ding{56} & 208.76 & 85.49 & 37.03 & 43.009 & 3.37 & 0.90 & 7.63 & 0.516 \\
(2) & \ding{52} & \ding{56} & \ding{56} & 193.51 & 60.14 & 24.03 & 30.970 & 2.40 & 0.81 & 4.11 & 0.344 \\
(3) & \ding{56} & \ding{52} & \ding{56} & 30.94 & 10.34 & 19.40 & 13.291 & 3.20 & 0.83 & 7.43 & 0.348 \\
(4) & \ding{52} & \ding{52} & \ding{56} & \underline{4.51} & \underline{1.57} & \textbf{8.40} & \textbf{0.590} & \underline{2.34} & \underline{0.79} & \underline{4.10} & \underline{0.337} \\
(5) & \ding{52} & \ding{52} & \ding{52} & \textbf{3.64} & \textbf{1.18} & \underline{10.03} & \underline{0.680} & \textbf{1.25} & \textbf{0.52} & \textbf{1.58} & \textbf{0.220} \\
\hline
\end{tabular}}
\vspace{-15pt}
\end{table}

\textbf{Impact of Dynamic Batching and V\&R:} Dynamic batching alone provides moderate improvement by ensuring spatial overlap, but fails to prevent catastrophic local failures on HAV (ATE drops only from 208.76 to 193.51~m). In contrast, V\&R is the primary driver of robustness; even without dynamic batching, it reduces HAV ATE to 30.94~m by actively rejecting inconsistent references. Their combination yields a synergistic effect, plummeting the ATE to 4.51~m. This confirms that constructing geometrically consistent contexts (Dyn.) and explicitly gating reliability (V\&R) are complementary and both indispensable.

\textbf{Role of Pose Graph Optimization:} PGO further reduces the absolute ATE to 3.64~m on HAV and from 2.34 to 1.25~m on Residence. Interestingly, on HAV, while ATE and ARE improve, RTE and RRE increase slightly. This is a theoretically expected behavior in SLAM: PGO optimizes the global trajectory to minimize absolute drift (ATE/ARE) by distributing errors across the graph, which can occasionally slightly compromise local relative consistency (RTE/RRE) if the local VFM predictions were already highly coherent. Nevertheless, the overall global geometric consistency is significantly enhanced, as evidenced by the uniform metric improvements on the longer Residence scene.

\section{CONCLUSION}
\label{sec:conclusion}

In this paper, we introduced \textit{On-the-Fly3R}, a training-free progressive reconstruction framework designed to scale feed-forward 3D visual foundation models to large-scale, unordered UAV imagery. Our approach addresses the memory bottlenecks and strict spatiotemporal assumptions of existing methods through two key components: retrieval-guided dynamic subset construction for geometrically consistent local context, and a validation-rejection-retry mechanism to actively suppress cascading error propagation. Extensive evaluations across diverse indoor and outdoor benchmarks confirm that \textit{On-the-Fly3R} can process thousands of images and square-kilometer UAV scenes within a bounded GPU memory budget, matching the accuracy of offline global inference. Moving forward, we plan to investigate semantic-assisted retrieval for challenging textures and explore fully incremental backend optimization to enable true real-time dynamic scene reconstruction.










\bibliographystyle{IEEEtran}
\bibliography{ref} 

@inproceedings{schonberger2016colmap,
  title     = {Structure-from-Motion Revisited},
  author    = {Sch{\"o}nberger, Johannes L. and Frahm, Jan-Michael},
  booktitle = {Proceedings of the IEEE Conference on Computer Vision and Pattern Recognition (CVPR)},
  pages     = {4104--4113},
  year      = {2016},
  doi       = {10.1109/CVPR.2016.445}
}

@article{zhan2025ontheflysfm,
  title   = {SfM on-the-fly: A robust near real-time SfM for spatiotemporally disordered high-resolution imagery from multiple agents},
  author  = {Zhan, Zongqian and Yu, Yifei and Xia, Rui and others},
  journal = {ISPRS Journal of  Photogrammetry and Remote Sensing },
  volume  = {224},
  pages   = {202--221},
  year    = {2025},
}

@article{zhan2026ontheflysfm,
  title   = {On-the-Fly Feedback Structure From Motion: Online explore-and-exploit unmanned aerial vehicle photogrammetry with incremental mesh quality-aware indicator and predictive path planning},
  author  = {Lou, Liyuan and Li, Wanyun and Gan, Wentian and others},
  journal = {IEEE Geoscience and Remote Sensing Magazine},
  volume  = {14},
  pages   = {219--239},
  year    = {2026},
}

@article{ATDOM,
  title   = {A-TDOM: Active TDOM via On-the-Fly 3DGS},
  author  = {Xu, Yiwei and Wang, Xiang and Yu, Yifei and others},
  journal = {International Journal of Computer Vision},
  volume  = {134},
  year    = {2026},
}

@inproceedings{wang2024dust3r,
  title     = {{DUSt3R}: Geometric {3D} Vision Made Easy},
  author    = {Wang, Shuzhe and Leroy, Vincent and Cabon, Yohann and others},
  booktitle = {Proceedings of the IEEE/CVF Conference on Computer Vision and Pattern Recognition (CVPR)},
  pages     = {20697--20709},
  year      = {2024},
  doi       = {10.1109/CVPR52733.2024.01956}
}

@inproceedings{leroy2024mast3r,
  title     = {Grounding Image Matching in {3D} with {MASt3R}},
  author    = {Leroy, Vincent and Cabon, Yohann and Revaud, Jerome},
  booktitle = {European Conference on Computer Vision (ECCV)},
  pages     = {71--91},
  year      = {2024},
  publisher = {Springer},
  doi       = {10.1007/978-3-031-73220-1_5}
}

@inproceedings{wang2025vggt,
  title     = {{VGGT}: Visual Geometry Grounded Transformer},
  author    = {Wang, Jianyuan and Chen, Minghao and Karaev, Nikita and others},
  booktitle = {Proceedings of the IEEE/CVF Conference on Computer Vision and Pattern Recognition (CVPR)},
  pages     = {5294--5306},
  year      = {2025},
  doi       = {10.1109/CVPR52734.2025.00499}
}

@inproceedings{yang2025fast3r,
  title     = {{Fast3R}: Towards {3D} Reconstruction of 1000+ Images in One Forward Pass},
  author    = {Yang, Jianing and Sax, Alexander and Liang, Kevin J. and others},
  booktitle = {Proceedings of the IEEE/CVF Conference on Computer Vision and Pattern Recognition (CVPR)},
  pages     = {21924--21935},
  year      = {2025},
  doi       = {10.1109/CVPR52734.2025.02042}
}

@inproceedings{wang2026pi3,
  title     = {{$\pi^3$}: Permutation-Equivariant Visual Geometry Learning},
  author    = {Wang, Yifan and Zhou, Jianjun and Zhu, Haoyi and others},
  booktitle = {International Conference on Learning Representations (ICLR)},
  year      = {2026}
}

@inproceedings{keetha2026mapanything,
  title        = {{MapAnything}: Universal Feed-Forward Metric {3D} Reconstruction},
  author       = {Keetha, Nikhil and M{\"u}ller, Norman and Sch{\"o}nberger, Johannes and others},
  booktitle    = {International Conference on 3D Vision (3DV)},
  pages        = {499--509},
  year         = {2026},
  organization = {IEEE},
  doi          = {10.1109/3DV69130.2026.00054}
}

@inproceedings{wang2026vggtomega,
  title     = {{VGGT-$\Omega$}},
  author    = {Wang, Jianyuan and Chen, Minghao and Zhang, Shangzhan and others},
  booktitle = {Proceedings of the IEEE/CVF Conference on Computer Vision and Pattern Recognition (CVPR)},
  pages     = {21486--21499},
  year      = {2026}
}

@inproceedings{wang2025spann3r,
  title        = {{3D} Reconstruction with Spatial Memory},
  author       = {Wang, Hengyi and Agapito, Lourdes},
  booktitle    = {2025 International Conference on 3D Vision (3DV)},
  pages        = {78--89},
  year         = {2025},
  organization = {IEEE},
  doi          = {10.1109/3DV66043.2025.00013}
}

@inproceedings{wang2025cut3r,
  title     = {Continuous {3D} Perception Model with Persistent State},
  author    = {Wang, Qianqian and Zhang, Yifei and Holynski, Aleksander and others},
  booktitle = {Proceedings of the IEEE/CVF Conference on Computer Vision and Pattern Recognition (CVPR)},
  pages     = {10510--10522},
  year      = {2025},
  doi       = {10.1109/CVPR52734.2025.00983}
}

@inproceedings{zhuo2025streamvggt,
  title     = {Streaming Visual Geometry Transformer},
  author    = {Zhuo, Dong and Zheng, Wenzhao and Guo, Jiahe and others},
  booktitle = {International Conference on Learning Representations (ICLR)},
  year      = {2026}
}

@inproceedings{lan2026stream3r,
  title     = {{STream3R}: Scalable Sequential {3D} Reconstruction with Causal Transformer},
  author    = {Lan, Yushi and Luo, Yihang and Hong, Fangzhou and others},
  booktitle = {International Conference on Learning Representations (ICLR)},
  year      = {2026}
}

@inproceedings{chen2026ttt3r,
  title     = {{TTT3R}: {3D} Reconstruction as Test-Time Training},
  author    = {Chen, Xingyu and Chen, Yue and Xiu, Yuliang and others},
  booktitle = {International Conference on Learning Representations (ICLR)},
  year      = {2026}
}

@article{yuan2026infinitevggt,
  title   = {{InfiniteVGGT}: Visual Geometry Grounded Transformer for Endless Streams},
  author  = {Yuan, Shuai and Yang, Yantai and Yang, Xiaotian and others},
  journal = {arXiv preprint arXiv:2601.02281},
  year    = {2026}
}

@article{chen2026lingbotmap,
  title   = {Geometric Context Transformer for Streaming {3D} Reconstruction},
  author  = {Chen, Lin-Zhuo and Gao, Jian and Chen, Yihang and others},
  journal = {arXiv preprint arXiv:2604.14141},
  year    = {2026}
}

@inproceedings{deng2025vggtlong,
  title     = {{VGGT-Long}: Chunk It, Loop It, Align It---Pushing {VGGT}'s Limits on Kilometer-Scale Long {RGB} Sequences},
  author    = {Deng, Kai and Ti, Zexin and Xu, Jiawei and others},
  booktitle = {IEEE International Conference on Robotics and Automation (ICRA)},
  year      = {2026}
}

@inproceedings{maggio2025vggtslam,
  title     = {{VGGT-SLAM}: Dense {RGB} {SLAM} Optimized on the {SL(4)} Manifold},
  author    = {Maggio, Dominic and Lim, Hyungtae and Carlone, Luca},
  booktitle = {Advances in Neural Information Processing Systems (NeurIPS)},
  volume    = {38},
  year      = {2025},
  doi       = {10.52202/085713-4324}
}

@inproceedings{zhang2026talo,
  title     = {{TALO}: Pushing {3D} Vision Foundation Models Towards Globally Consistent Online Reconstruction},
  author    = {Zhang, Fengyi and Zhang, Tianjun and Khosoussi, Kasra and others},
  booktitle = {Proceedings of the IEEE/CVF Conference on Computer Vision and Pattern Recognition (CVPR)},
  pages     = {21870--21879},
  year      = {2026}
}

@inproceedings{xie2026scal3r,
  title     = {{Scal3R}: Scalable Test-Time Training for Large-Scale {3D} Reconstruction},
  author    = {Xie, Tao and Yang, Peishan and Jin, Yudong and others},
  booktitle = {Proceedings of the IEEE/CVF Conference on Computer Vision and Pattern Recognition (CVPR)},
  pages     = {21760--21771},
  year      = {2026}
}

@inproceedings{cheng2026merg3r,
  title     = {{MERG3R}: A Divide-and-Conquer Approach to Large-Scale Neural Visual Geometry},
  author    = {Cheng, Leo Kaixuan and Shaikh, Abdus and Liang, Ruofan and others},
  booktitle = {Proceedings of the IEEE/CVF Conference on Computer Vision and Pattern Recognition (CVPR)},
  pages     = {28969--28978},
  year      = {2026}
}

@inproceedings{shen2025fastvggt,
  title     = {{FastVGGT}: Fast Visual Geometry Transformer},
  author    = {Shen, You and Zhang, Zhipeng and Qu, Yansong and others},
  booktitle = {International Conference on Learning Representations (ICLR)},
  year      = {2026}
}

@article{shi2026supscene,
  title   = {{SupScene}: Scene-Structured Overlap Supervision for Image Retrieval in Unconstrained {SfM}},
  author  = {Shi, Xulei and Wang, Maoyu and Peng, Yuning and others},
  journal = {arXiv preprint arXiv:2601.11930},
  year    = {2026}
}

@article{xiong2024gauuscenev2,
  title   = {{GauU-Scene V2}: Assessing the Reliability of Image-Based Metrics with Expansive {LiDAR} Image Dataset Using {3DGS} and {NeRF}},
  author  = {Xiong, Butian and Zheng, Nanjun and Liu, Junhua and others},
  journal = {arXiv preprint arXiv:2404.04880},
  year    = {2024}
}

@inproceedings{lin2022urbanscene3d,
  title     = {Capturing, Reconstructing, and Simulating: The {UrbanScene3D} Dataset},
  author    = {Lin, Liqiang and Liu, Yilin and Hu, Yue and others},
  booktitle = {European Conference on Computer Vision (ECCV)},
  pages     = {93--109},
  year      = {2022},
  publisher = {Springer},
  doi       = {10.1007/978-3-031-20074-8_6}
}

@inproceedings{shotton2013scene,
  title     = {Scene Coordinate Regression Forests for Camera Relocalization in {RGB-D} Images},
  author    = {Shotton, Jamie and Glocker, Ben and Zach, Christopher and others},
  booktitle = {Proceedings of the IEEE Conference on Computer Vision and Pattern Recognition (CVPR)},
  pages     = {2930--2937},
  year      = {2013},
  doi       = {10.1109/CVPR.2013.377}
}
\end{document}